\documentclass{article}
\usepackage[final,nonatbib]{nips_2016}\usepackage[utf8]{inputenc}\usepackage[T1]{fontenc}
\usepackage{hyperref,url,booktabs,amsfonts,nicefrac,microtype,subfig,graphicx}
\usepackage{caption}
\usepackage[table]{xcolor}
\graphicspath{{./}}

\title{A Growing Long-term Episodic \& Semantic Memory}
\author{
  Marc Pickett, Rami Al-Rfou, Louis Shao, Chris Tar %, Javier Snaider, Matthew Henderson,
  \\ Google Research, Mountain View, CA, USA\\
\texttt{pickett,rmyeid,overmind,ctar@google.com}\\
}

\begin{document} \maketitle %((((

% ; Abstract (4 sentences):
\begin{abstract} %(
  % 1. State the problem
  The long-term memory of most connectionist systems lies entirely in the weights of the
  system.  Since the number of weights is typically fixed, this bounds the total amount of
  knowledge that can be learned and stored.
  % 2. Say why it's an interesting problem
  Though this is not normally a problem for a neural network designed for a specific task, such
  a bound is undesirable for a system that continually learns over an open range of domains.
  % 3. Say what your solution achieves
  To address this, we describe a lifelong learning system that leverages a fast, though
  non-differentiable, content-addressable memory which can be exploited to encode both a long
  history of sequential episodic knowledge and semantic knowledge over many episodes for an
  unbounded number of domains.
  % 4. Say what follows from your solution
  This opens the door for investigation into transfer learning, and leveraging prior knowledge
  that has been learned over a lifetime of experiences to new domains.
\end{abstract} %)

% Abstracts should be at most 4 pages long (not including references).  Supplementary materials
% can be added with more details about the proposed approach (no page limit), but the reviewers
% will not be required to go over them.

\section{Introduction} %)))(((
% ; Introduction (1 page):
% 1. Describe the problem (example)
% 2. State your contributions (bullets, refutable)
% 3. ... and that is all

% Motivation:
Over the course of several decades of experience a person typically learns a variety of
disparate domains.  A person can learn a new domain and still retain long-held knowledge about
previously learned domains.  However, many neural models experience {\em catastrophic
  interference} when being trained on new domains, where the new learning overrides or corrupts
the network's earlier knowledge.
If a machine were similarly capable of learning a variety of domains over its ``lifetime'', it
would potentially allow the machine to transfer knowledge among domains and bring to bear a
large box of tools when facing a new problem.
The problem we address is: {\em How can a machine store an unbounded amount of episodic and
  semantic memory such that it can store knowledge from previous tasks and efficiently retrieve
  relevant information for new tasks?}
To do this, we must also address the subproblems of how a machine can automatically segment its
experiences into episodes and how it can encode episodes into long-term memory such that
relevant semantic knowledge and analogous episodes can be efficiently recalled and applied to
new experiences.

Our current work makes the following contributions:
1. We introduce a system that stores both episodic and semantic memory in a single memory
  system\footnote{We take a liberal definition of ``episodic memory'', which includes memory
    of specific sequential events.  This is a looser definition than that used by Tulving and
    others \cite{tulving1993episodic}, who require that episodic memory be autobiographical,
    for example.  We define ``semantic memory'' loosely as abstractions or summaries induced
    over multiple sequences.}.
2. This system automatically separates domains without requiring an explicit signal telling
  it which domain it is currently experiencing.
  We describe how a classifier may be used with this system so that it may retrieve relevant
  domain information while only explicitly considering a fraction of its knowledge of previous
  domains.  This is in contrast to other systems that either require explicit domain indicators
  or linearly consider each of its known domains in turn.
3. We describe how the system may be used to automatically increase its memory capacity and
  overcome catastrophic interference.

\section{Lifelong Unsupervised Learning} %)))(((
% ; Problem (1 page): (background, whiteboard)
By any reasonable measure, human brains well over several trillion parameters\footnote{A
  healthy adult cortex is estimated to have roughly 20 billion neurons and 150 trillion
  synapses \cite{drachman2005we}.  The estimates for the number of bits captured by these
  synaptic connections vary widely.  A recent estimate gives 4.7 bits per synapse
  \cite{bartol2015nanoconnectomic}, yielding roughly 700 trillion bits.  At 32 bits per
  floating point parameter, this gives roughly 5 trillion floating point parameters in the
  cortex.  By contrast, it is currently rare for artificial neural networks to have more than
  100 billion floating point parameters \cite{trask2015modeling}, with typical networks having
  much fewer.  For example, a recent ResNet architecture for CIFAR-10 had only 1.7 million
  parameters \cite{he2015deep}.}.  If an artificial neural network is going to reach
human-level intelligence, it seems likely that it will also need a similar order of magnitude
of parameters.
Assuming typical bounds on each parameter (e.g., 32 bits), for a machine to represent a huge
amount of knowledge about the world, it will need a large number of parameters.
% Fixed brain
One way to achieve this, which we call a {\em fixed-brain} design, is for the machine to begin
its existence with nearly all the parameters it will ever have.  This a perfectly reasonable
approach, as there is evidence that the total number of neurons in humans actually {\em
  decreases} with age, even accounting for neurogenesis \cite{pakkenberg1997neocortical}.
However, a fixed-brain design places an upper bound on the machine's total domain knowledge,
which requires the machine's designers to have some prior knowledge of an upper bound of the
complexity of the machine's lifetime experience (which has the potential to be orders of
magnitude greater than the lifetime experience of a single person).
% Assuming a connectionist architecture, it also requires that the designers either specify the
% topology for a huge netowrk, or a way for the topolgy to constructed automatically.  Finally,
% a fixed-brain design has the drawback that its designers will be debugging a system that
% starts with billions of parameters.

% Growing brain
Instead, we are investigating an alternative to a fixed-brain design we call a {\em
  growing-brain} design, where a machine has the ability to indefinitely allocate (and
deallocate) new parameters as needed from an extendable memory.
At the heart of our approach we assume an unlimited associative memory to which our system can
{\em read} and {\em write} real-valued vectors of some fixed width\footnote{Note that this
  memory is still technically bounded by its ``address space'', but this is exponential in the
  vector width, and is unbounded for all practical purposes.
  % so this bound can feasibly by made to be sufficient to pinpoint any time and location in
  % the history of the universe (i.e., 820 bits can designate any of the fewer than $10^{247}$
  % Plank voxels at any of the 13 billion years worth of Plank times).
} (e.g., 10,000 elements).  We assume each of these operations takes time that is logarithmic
in the number of items in the memory.  A ``write'' operation takes a key value pair, where both
the {\em key} and {\em value} are vectors, and simply stores them in memory.
A ``read'' operation takes a {\em key} vector and returns a small set of vectors that are
likely to be those whose keys are closest to the given key.
In the absence of other information, this memory assumes that nearby keys map to nearby values.
So, unlike a normal hash table, a key during retrieval need not exactly match the key that was
used to originally store the value.  Note that a vector can serve as its own key, which results
in {\em content addressable memory}.  That is, one can read from the memory using a noisy or
incomplete version of a vector and retrieve a completed denoised version.
Various models have been proposed for this type of memory, such as Sparse Distributed Memory
\cite{Kanerva:1988:SDM:534853}, Clean-up Memory \cite{stewart2011biologically}, and approximate
nearest neighbor search methods \cite{wang2014hashing}.
Few of these methods are differentiable, and we sacrifice the assumption of differentiability
in our vector memory, which gives us flexibility in which systems we can use.

% Universal-LSTM Principles
We make the following assumptions in our approach:
% \begin{itemize} %(
% \item We want to understand War&Peace (& many other stories)
% \item Same fundamental mechanism for mazes as all other structures (Buzsaki hypothesis)
1. The memory capacity of a single vector in memory has a fixed bound that is less than the
  amount of information we eventually want to encode about the world.
2. Following \cite{schmidhuber2015learning}, our system is unsupervised, and its goal is to
  compress its experiences, which is an uninterrupted stream of fixed width vectors.  This
  includes both episodic knowledge (individual instances of sequences) and semantic knowledge
  (patterns among many sequences).
3. Nearly all knowledge learned about the world is stored in the associative memory.  This
  includes both individual episodes and semantic knowledge.  We allow a fixed number of learned
  parameters in a meta-level controller outside of the vector memory.
4. Operations on the vector memory, such as insertion, deletion, and retrieval, are not
  assumed to be differentiable.
% \end{itemize} %)

Motivated by very early infant development, we assume an unsupervised setup where our machine
experiences a continuous stream of data, but has no external supervision or reward signals, and
no actions to affect the environment.  The machine receives a continuous stream of fixed-width
vectors.  In our experiments we use the sequence of 1024-bit memory states from Atari games
concatenated with an 18-bit one-hot encoding of the previous action.  We use the implementation
available from OpenAI Gym \cite{openaigym}.  An example of this data is shown in Figure
\ref{figure:atariRAM} in the Supplementary Material.  Since our system has no control, it's
merely watching another (random) player play games.  Although our stream comes from multiple
runs from different Atari games, the machine is given no special signal marking the beginning
of an episode, nor is it given explicit information about which game is being played at any
time.  Though the machine's goal is merely to remember and compress these sequences, we
hypothesize that, in doing so, the machine will develop a model of the games that will be
useful for a later time when it is given a reward signal.

In this setup, we want our system to be capable of learning new games indefinitely without
forgetting earlier games.  We would also like the system to leverage knowledge from earlier
games to learn faster on new games.

\section{Solution Overview: Storing Program Vectors in Long-term Memory} %)))(((
% ; Idea (2 pages): (Intuition)

We assume our vector memory works on floating-point vectors of a fixed size (we somewhat
arbitrarily chose 64 elements for our implementations).  We now discuss how such a memory can
be used to store a virtually unlimited amount of sequential trace data.

% \subsection{Storing Thought Vectors from a Single LSTM} %))((

Inspired by Complementary Learning Systems \cite{o2014complementary}, we assume we have a
large, though fixed, memory buffer (separate from the extendable vector memory) in which we can
rotely store a long sequence of vectors.
Our initial approach for compressing the data in this buffer was to train an LSTM Sequence to
Sequence auto-encoder \cite{sutskever2014sequence} to encode subsequences from this longer
sequence (we used subsequences of length 7), then commit the 64-element-wide thought vectors
for the subsequences to the vector-memory.
There are several problems with this approach.  One is that the amount of {\em semantic}
knowledge is bounded by the weights of the LSTM auto-encoder.  This means that we cannot expect
the LSTM to keep learning the dynamics of new Atari games indefinitely.

% \subsection{Storing Thought Vectors from Multiple LSTMs} %))((

To get around this, the number of free parameters for the LSTM models needs to be increased.
One possibility was to simply increasing the number of hidden states in the LSTM, but each
expansion would require us to address how to train the grown LSTM without causing catastrophic
interference.

Our next approach was to train {\em multiple} LSTM auto-encoders, where each auto-encoder
attempted to compress the input then reported a loss based on the difference between the
original and decoded sequences.  For each sequence, we tied together the losses with a {\em
  minimum} operation, which had the effect of only training the model that best encoded the
sequence.  In our experiments, this caused the models to specialize: When we trained three
models on data from three different Atari games, each model specialized on encoding a
particular game.  (Of course, a single large LSTM with the same number of parameters has a
lower reconstruction error than three small LSTMs, but the latter approach has the advantage of
a simple straightforward way to extend the capacity of the model without risking catastrophic
interference.)

% \subsection{LSTM Embeddings using a ``Stretcher'' Network} %))((

%
One issue with using multiple LSTM auto-encoders is that our model's semantic knowledge is
stored outside the vector memory (i.e., in the weights of the LSTM auto-encoders).  Since each
model has 651,154 parameters, this is far too big to fit into a single vector of our vector
memory (which we chose to store vectors of 64 elements).  To address this, we reduced the
dimensionality of the auto-encoders in a manner reminiscent of HyperNetworks \cite{haHypernets}
and learnets \cite{DBLP:journals/corr/BertinettoHVTV16}.  We trained 64-element embedding
vectors for each LSTM by using a feedforward ``stretcher'' network shown in Figure
\ref{figure:singleProg} in the Supplementary Material, that ``stretches'' a vector of size 64
to size 651,154 using layers of 64, 128, 256, and 651,154 nodes.  The final weights of the
stretcher network are ``reshaped'' into the weights for an 64-hidden-unit LSTM auto-encoder.
All the layers of the stretcher network are fully connected except the last layer, which is
sparsely connected with only 1\% of the possible connections, chosen randomly (a fully
connected matrix would be too big to easily train).  Thus, the parameter specification of each
LSTM auto-encoder is a differentiable function of its 64-element embedding and the weights of
the stretcher network, and thus backpropagation adjusts the embedding and the weights of the
stretcher network instead of directly changing the auto-encoder's parameters.

%% [I originally called this an ``expander net'', but I didn't want to confuse this with the idea
%%   of our expanding memory.  I also like Network of Procrustes, but I wanted to be clear for the
%%   95\% of readers who didn't geek out on Greek mythology.]

We dub the final embedding for each LSTM auto-encoder a ``program vector'', with the analogy
that this embedding can be interpreted as a program that can be ``called'' with different
thought-vectors or ``arguments'' to produce specific sequences.
Of course, knowledge is stored in the weights of the stretcher network, which has a fixed
number of parameters.  We hope that the stretcher network becomes somewhat generic, only
encoding very general knowledge after training on a wide variety of
games\footnote{Alternatively, one can imagine using a meta-stretcher network that allows us to
  embed many different stretcher networks, which could add another level of generality.  At
  some point, there will have to be a fixed controller.}.

%% We find the following analogy useful: An Atari 2600 is a general console that can take a
%% specific Read Only Memory (ROM) for each game (these are 32,768 bits each), and stores its
%% state in its RAM (1024 bits).  In our analogy, the stretcher network is like the Atari console,
%% which takes program vectors as its ROMs, and produces a sequence of memory states,
%% corresponding to the hidden values of the LSTM.  A parallel can be made to the neural
%% programmer interpreter \cite{reed2015neural}, which learns a general, possibly Turing complete,
%% programming language.

We hypothesize that the set of practically useful LSTM auto-encoders is only a tiny fraction of
the set of those possible, and that the stretcher network will learn to generate ``sensible''
LSTMs with most 64-element vectors chosen from a 0, 1, Gaussian distribution.

% Put variational loss on Program vectors, so that random vectors fill the space of programs?

% - 99.9% of smarts are stored in LTM as both: vignettes (instances) and abstractions of vignettes (schemas and Von Neumann idea)
%   (Thought vectors don't capture learned correlations across many examples.  These are stored in the LSTM weights, so must also be in LTM.)
% - Remaining .1% is meta control system, like Neural Programmer Interpreter
% - Neurons are usually *off*
% - LTM is *expanding* and virtually unlimited.  Only constraint is ability to quickly retrieve relevant vectors (episodes and schemas) during parse.

% \section{The Not-Actually-Universal LSTM} %)))(((
% ; Details (5 pages):

% Not-Actually-Universal LSTM

\section{Preliminary Results} %)))(((

We trained the stretcher network by training on sequences from various Atari games and varying
the number of program vectors the system is allowed to use.  The strongest result we have so
far is that when the number of program vectors is equal to the number of Atari games, the
system tends to use the same program vector for the same domain.  That is, to some extent, it
automatically segments the domains without being given explicit information which Atari game
the traces are coming from.  (See Figure \ref{figure:domains} in the Supplementary Material).
% The system was able to express sequences of 7 by 1024 bits as vectors of size 64.

\section{Related work} %)))(((
% ; Related work (1-2 pages):

Many expandable architectures have been proposed both recently and several decades ago.
Non-parametric methods, such as Case-based reasoning \cite{aha1991instance}, K-means and others
(see \cite{hollander2013nonparametric} for a survey), have the ability to grow their capacity
linearly with the data.  Most of these methods operate on static vector data, so must be
adapted to operate on sequential data.  Furthermore, semantic knowledge (i.e., knowledge of
patterns) is usually stored only implicitly (i.e., in the data points), unlike our proposed
system which stores both instances and embeddings of semantic knowledge.

Other methods have been proposed to address catastrophic interference.  For example,
Complementary Learning Systems \cite{o2014complementary} and Learning without Forgetting
\cite{li2016learning} both interleave training of remembered earlier data with new data.  We
draw inspiration from both of these systems, and from work on Progressive Networks
\cite{rusu2016progressive}, which freezes weights of networks trained on earlier domains.
Unlike the others, Progressive Networks allow a network to expand its capacity.  Unlike our
system, Progressive Networks do not attempt to store the semantic knowledge of earlier systems
in a content addressable memory, and have the problem that their network grows quadratically in
the number of domains.  We hypothesize that storing semantic knowledge in a content addressable
memory will help address this by allowing fast lookup of relevant ``program vectors''
potentially yielding linear storage and logarithmic program lookup.

Several methods have been proposed for expanding the capacity of neural networks.  Part of our
work was initially inspired by the Cascade Correlation algorithm \cite{fahlman1989cascade}, an
early example which incrementally learns new features and adds them to a feed-forward network
while freezing weights for previously learned features.  Other models have since built on these
ideas such as growing neural gas \cite{fritzke1995growing}, Net2Net \cite{chen2015net2net}, and
AdaNet \cite{cortes2016adanet}.  Our work attempts to build on these ideas by providing a means
of storing sequential {\em instances} in addition to semantic (weight) information.

There have been recent advances in differentiable memory, such as Neural Turing Machines
\cite{graves2014neural}, Memory Networks \cite{weston2014memory, kumar2015ask}, Differentiable
Neural Computers \cite{gravesNature2016}, and Memory-based Deep Reinforcement Learning
\cite{oh2016control}.  All of these provide the system with what is essentially a {\em working
  memory} that can be accessed during the course of a single episode.  Unlike our system, the
memory is cleared between episodes, so the only {\em long-term} memory these systems retain is
in the network weights.

Episodic memory has also been a component of many cognitive architectures, such as SOAR
\cite{derbinsky2009efficiently}, LIDA \cite{ramamurthy2011memory}, and CLARION
\cite{sun2012memory}.  Our work was originally influenced specifically by SOAR
\cite{derbinsky2009efficiently}, but extends these by using recent developments in sequence to
sequence models to encode sequences as static vectors.
Episodic memory has been shown to be useful for Reinforcement Learning tasks
\cite{blundell2016model}.  Our system provides a mechanism by which episodes may be stored and
retrieved.

%% We view the program vectors as a general way of encoding specialized modules that are capable
%% of performing a range of functions.  Parallels can be made to reusable neural modules
%% \cite{reisinger2004evolving} such as cortical minicolumns \cite{hawkins+blakeslee:2004,
%%   mountcastle1978organizing}.

\section{Open Challenges and Future work} %)))(((
% ; Conclusions and further work (0.5 pages):

The primary contribution of our work is a system for encoding an unbounded amount of episodic
and semantic knowledge in a expandable content-addressable vector memory.
This work is still in its infancy and there are many unresolved issues to answer the question
of how a machine can store a lifetime of knowledge such that it can be usefully retrieved and
transferred to new situations.
We share our current approaches for addressing some of these problems in the Supplementary
Material.

\bibliographystyle{amsplain}\bibliography{marcbib} %)))(((

\section*{Supplementary Material} %)))(((

\begin{figure}[h] %(
  \centering {\includegraphics[height=60mm]{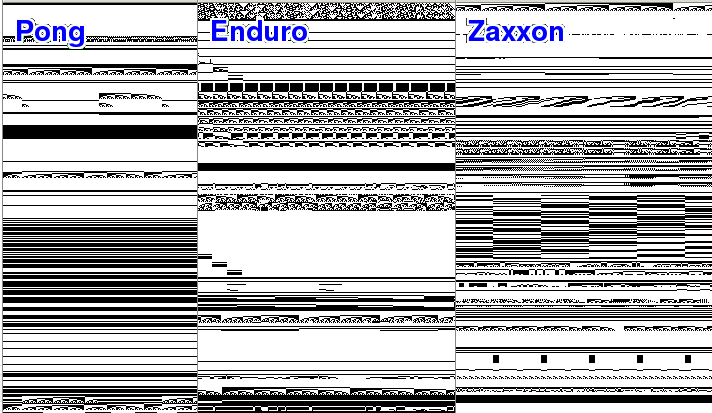}} {\caption{{\bf Atari RAM states over
        time.} Shown above is part of the memory states for traces collected from Atari Pong,
      Enduro, and Zaxxon.  The x-axis is time steps, and the y-axis is the first 400 (of 1024 +
      18) RAM-bits for each game, where white represents 0 and black represents 1, and the
      lowest index at the top.}
    \label{figure:atariRAM}}
\end{figure} %)

\begin{figure} %(
  \centering
  \subfloat[A Stretcher Network]{{\includegraphics[height=50mm]{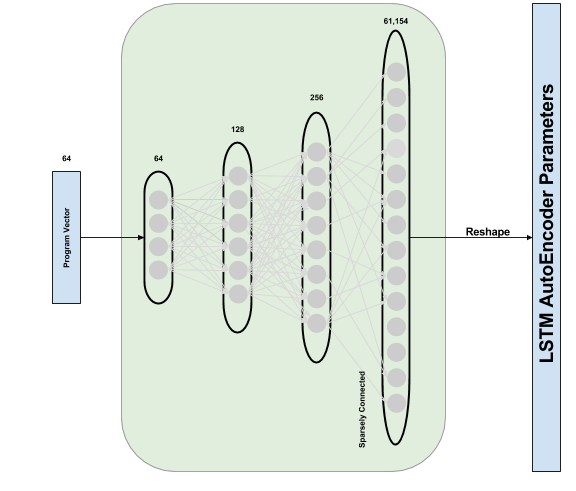}} \label{figure:singleProg}}
  \qquad
  \subfloat[A Stretcher Network used Three Times]{{\includegraphics[height=50mm]{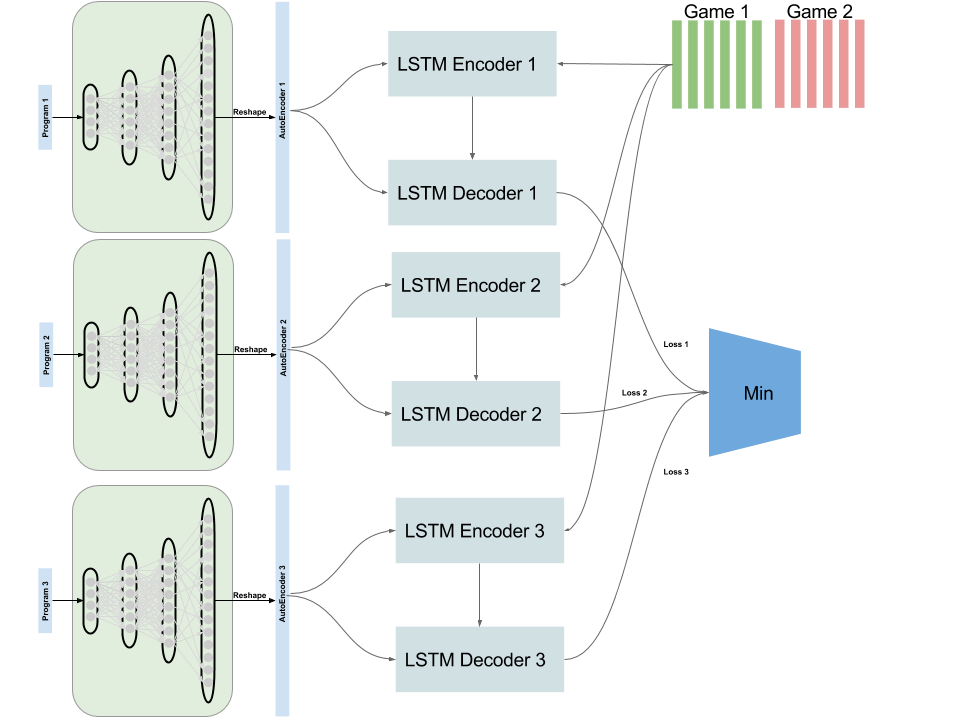}} \label{figure:tripleProg}}
  \caption{{\bf Stretcher Networks} Figure \ref{figure:singleProg} shows a single stretcher
    network, that takes a relatively small vector and produces a much larger vector, which is
    then reshaped into the parameters of an LSTM auto-encoder.  Figure \ref{figure:tripleProg}
    shows the same network used three times, producing three different LSTM autoencoders from
    three different ``program'' vectors.  Note that the weights of the stretcher network are
    tied for all three instances.}
  \label{figure:singletriple}
\end{figure} %)

\begin{figure} %(
  \centering {\includegraphics[height=80mm]{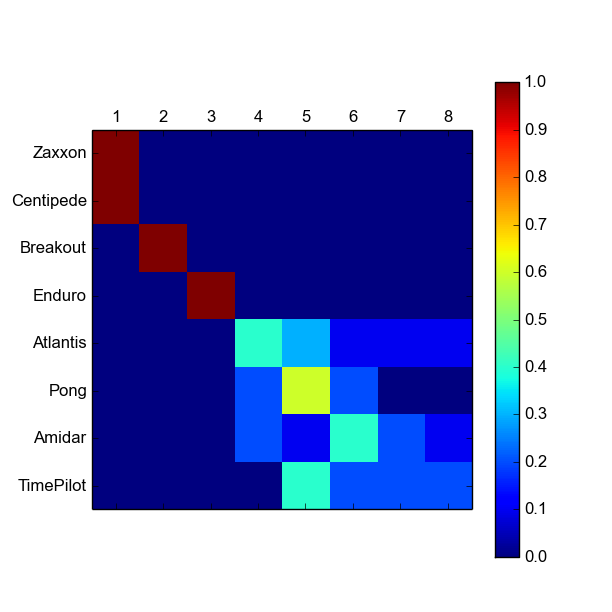}} {\caption{{\bf Usage of Program Vectors
        by Domain} This plot shows the usage of program vectors by traces collected from
      different domains.  For example, the system used program vector 1 consistently for traces
      collected from Zaxxon or Centipede.  This is without being given explicit information
      about which Atari game traces came from.  Note that the program vectors are arbitrarily
      indexed.}
    \label{figure:domains}}
\end{figure} %)

\section{Approaches for Addressing Shortcomings of the Current Approach} %)))(((

Below, we provide a sample of the shortcomings of the current approach and what we are doing to
address them.

\subsection{Train simple seq2vec classifier that produces best program vector for input vector} %))((

% How to recall a subset of relevant program vectors?  Classifier and learn keys
When our current system is given a trace, it checks every program vector in its vector memory,
one by one, and encodes the traces using the program vector with the smallest loss.  This
linear search is undesirable when there are thousands or millions of program vectors, which a
continually learning system might accumulate over a lifetime.
Instead, we propose a system that quickly retrieves a small subset of relevant program vectors
given a trace.  We do this by augmenting each program vector with another {\em key} vector.
When given an input key, the vector memory retrieves the vectors whose keys are closest to the
input key.  We then train a classifier that takes a trace as input and generates a key.  This
key is then fed into the vector memory, which generates a (small) set of candidate program
vectors.  The classifier's loss is the difference between the key it generates and the key for
the best program vector.  This puts domain knowledge into the classifier, which has a fixed
number of parameters.  To remedy this, we initially train both the classifier and the program
keys.  Once the classifier begins to stabilize (and we have a ``universal'' classifier), we
train only the program vectors' keys (so that these keys effectively contain an encoding of
{\em in what situations} to use the program vectors they point to).

\subsection{Train simple greedy growing number of LTM program vectors} %))((

Another direction is to automatically allocate new program vectors with experience.  The
simplest case of this is where we have a batch dataset.  In this case, we incrementally add
program vectors (initializing them to be nearby existing program vectors), then train with the
new vector.  We keep adding until the cost of storing the new vector is no longer offset by the
reduction in reconstruction error.

For the online case, we can compress the incoming sequence using our existing program vectors
until the buffer reaches some limit.  (This buffering can be done by the vector memory.)  When
the buffer is full (which will take longer as the system learns more patterns), the system can
add program vectors and train them on the data in the buffer, similar to the batch case, but
taking the buffer as the batch.  The system can also retrieve earlier memories from the vector
memory, and interleave these in training along with the data in the buffer, similar to the
Complementary Learning Systems model \cite{o2014complementary}.

\subsection{Make predictions via instance retrieval (use K-nearest neighbors to transfer knowledge)} %))((

Although the current system is an autoencoder, there are at least two minor modifications that
can turn it into a prediction model.  The most straightforward is to use prediction loss
instead of reconstruction loss while training the stretcher network and program vectors.

An alternate approach is more closely related to Case-base Reasoning.  This is where, for each
thought vector (with its program vector), we simply use the vector memory to memorize the next
{\em consequent} thought vector in the sequence.  When given a new thought-with-program vector,
we retrieve nearby vectors, and use the memorized consequent vectors to predict the consequent
vector for the input.  This could be a weighted average, or we could simply allow multiple
predictions.  This latter approach potentially allows for richer summaries of possible
predictions, such as a disjoint distribution over divergent predictions.

\subsection{Smarter parsing: Try multiple parsings} %))((

Our current model doesn't do a search for parsing the data stream.  It simply breaks the chunks
into sequences of a particular length.  The system might encode the stream more compactly if it
instead explores multiple windows on which to parse.  Although there have been methods for a
differentiable system to learn to parse \cite{chung2016hierarchical}, our first attempts at
this will be a simpler discrete search over possible parsings and segmentations.  This allows
us to easily add other functionality, such as top-down contextual influences of a hierarchical
system and classical parsing ideas using dynamic programming and back-tracking, to the parsing
process.

\subsection{Train ``continuation'' version, where sequences ``call'' next sequence} %))((

When we use the decoder to unroll thought vectors in our current model, we expect to get back
just the literal original (short) sequence used in encoding the thought vector.  There are
variations on this idea one may try.  Most obviously is prediction or skip-thought
\cite{kiros2015skip}, in which we try to {\em predict} the following sequence rather than
recite the current sequence.

Another approach we would like to investigate is the idea that an unrolled sequence can {\em
  call} other sequences.  Instead of generating only literal base-level vectors, a unrolled
sequence can contain a length-two subsequence corresponding to a program vector followed by its
argument (i.e., a thought vector).  One simple approach for training this would be to generate
the program vector and thought vector for the last subsequence of a long sequence, then extend
the previous subsequence with the short sequence containing these two vectors.  For example, if
our window size is 4, and we want to encode the sequence of vectors $A, B, C, D, E, F, G, H$,
then we first encode $E, F, G, H$.  Suppose the encoder encodes this sequence as program vector
$P_7$ using thought vector $\theta$.  Then we feed the system $A, B, C, D$, but ask it to
produce a sequence with {\em six} elements: $A, B, C, D, P_7, \theta$.  We can also extend this
idea so that sequences can call program vectors at other points rather than just pointing to
their continuation.

\subsection{Reusable Submodules, Mixture of Experts, and Explaining Away} %))((

% Mixture of experts and explaining away (submodules)

A pattern that is common across many Atari games is a binary ``counter'' in the first 8 bits of
memory\footnote{These counters (and other patterns) are common in other bit indices also, but
  we would like to address the simpler ``non-transformed'' case first.}.  However, in the
current system, each short trace is encoded by only one program vector, so each program vector
needs to independently represent this ``bit counter'' pattern.  Although the program vectors
implicitly share knowledge through the stretcher network, it would be useful if ``expert''
program vectors were able to specialize on patterns that are used across different domains.

One approach for addressing this is by incrementally ``explaining away'' elements of a trace by
greedily applying program vectors that most reduce the reconstruction cost.  For example, if a
trace has a binary counter pattern in its first 8 bits, and if we have a program vector $P_b$
that specializes in this pattern, then we can encode the thought vector created by encoding the
trace with program vector $P_b$.  Ideally, the decoding of this thought vector using $P_b$
would produce exactly the binary counter subsequence.  We would then subtract the decoded
sequence (each vector element-wise) from the original sequence (which would essentially cause
the first 8 bits of the vectors in the trace to all be near zero), and repeat until no program
vectors were able to further reduce the reconstruction cost.  We would then train the various
``expert'' program vectors based on the calls that this search made.

We could then store the sequence of calls to the experts the same way we store other sequences.
To create the original trace, we would decode the sequence of calls and allow each expert to
additively modify the ``canvas'' of the trace (where experts might ``add'' negative numbers) in
a manner reminiscent of DRAW \cite{gregor2015draw}.

\end{document}